\documentclass{article}

\usepackage{microtype}
\usepackage{graphicx}

\usepackage{subfig}
\usepackage{color}

\usepackage{hyperref}

\usepackage[preprint]{neurips_2019}
\usepackage{booktabs} 
\usepackage{amsmath}
\usepackage{amssymb}
\usepackage{multirow}
\usepackage{xspace}
\usepackage{algorithm}
\usepackage{algorithmic}

\usepackage{hyperref}

\usepackage[utf8]{inputenc} 
\usepackage[T1]{fontenc}    
\usepackage{hyperref}       
\usepackage{url}            
\usepackage{booktabs}       
\usepackage{amsfonts}       
\usepackage{nicefrac}       
\usepackage{microtype}      

\newcommand{\stopforward}{\texttt{sf}}
\newcommand{\stopgradient}{\texttt{sg}}
\newcommand{\petridishhard}{Isolated\xspace}
\newcommand{\petridishsoft}{Joint\xspace}
\newcommand{\Petridish}{Petridish\xspace}

\newcommand{\NASDIR}{.}

\DeclareMathOperator*{\argmin}{arg\,min}

\title{Efficient Forward Architecture Search}

\author{
Hanzhang Hu,\textsuperscript{\rm 1}
John Langford,\textsuperscript{\rm 2}
Rich Caruana,\textsuperscript{\rm 2} \\
{\bf Saurajit Mukherjee,\textsuperscript{\rm 2}
Eric Horvitz,\textsuperscript{\rm 2}
Debadeepta Dey\textsuperscript{\rm 2}}\\
\textsuperscript{\rm 1}Carnegie Mellon University,
\textsuperscript{\rm 2}Microsoft Research\\
hanzhang@cs.cmu.edu, \{jcl,rcaruana,saurajim,horvitz,dedey\}@microsoft.com
}
\begin{document}

\maketitle

\begin{abstract}
We propose a neural architecture search (NAS) algorithm, \Petridish, to iteratively 
add shortcut connections to existing network layers. The added shortcut connections 
effectively perform gradient boosting on the augmented layers.
The proposed algorithm is motivated by the feature selection algorithm 
forward stage-wise linear regression, since we consider NAS as a generalization 
of feature selection for regression, where NAS selects shortcuts among layers 
instead of selecting features. 
In order to reduce the number of trials of possible connection combinations, we train
jointly all possible connections at each stage of growth while leveraging
feature selection techniques to choose a subset of them. 
We experimentally show this process to be an efficient forward 
architecture search algorithm that can find competitive 
models using few GPU days in both the search space of repeatable 
network modules (cell-search) and the space of general networks (macro-search). 
\Petridish is particularly well-suited for warm-starting from existing models 
crucial for lifelong-learning scenarios.
\end{abstract}

\section{Introduction}

Neural networks have achieved state-of-the-art performance on large scale supervised learning tasks across domains like
computer vision, natural language processing, audio and
speech-related tasks using architectures manually designed by skilled
practitioners, often via trial and
error. Neural architecture search (NAS)~\citep{nas,NASCell,Real2018RegularizedEF,Pham2018EfficientNA,Liu2018DARTSDA,proxyless} algorithms attempt to automatically find good architectures given data-sets.
In this work, we view NAS as a bi-level combinatorial optimization problem ~\citep{Liu2018DARTSDA}, where we seek both the optimal architecture and its associated optimal parameters.  Interestingly, this formulation 
generalizes the well-studied problem of feature selection for linear regression~\citep{lasso,lars,omp_fr}. This observation permits us to draw and leverage parallels between NAS algorithms and feature selection algorithms. 

A plethora of NAS works have leveraged sampling methods
including reinforcement learning~\citep{nas,NASCell,Liu2018HierNA},
evolutionary
algorithms~\citep{Real2017EvoNet,Real2018RegularizedEF,Elsken2018EfficientMN},
and Bayesian optimization~\citep{Kandasamy2018BNAS} to enumerate architectures that are then independently trained. Interestingly, these approaches are uncommon for feature
selection.  Indeed, sample-based NAS often takes hundreds of GPU-days to find good architectures, and can be barely
better than random search~\citep{Elsken2018NeuralAS}.

Another common NAS approach is analogous to sparse optimization~\citep{lasso} or
backward elimination for feature selection,
e.g.,~\citep{Liu2018DARTSDA,Pham2018EfficientNA,proxyless,snas}.  The
approach starts with a super-graph that is the union of all possible
architectures, and learns to down-weight the unnecessary edges
gradually via gradient descent or reinforcement learning. Such
approaches drastically cut down the search time of NAS. However,
these methods require domain knowledge to create the initial super-graphs, and 
typically need to reboot the search if the domain knowledge is updated.


In this work, we instead take an approach that is analogous to a
forward feature selection algorithm and iteratively grow
existing networks. Although forward methods such as Orthogonal
Matching Pursuit~\citep{omp} and Least-angle Regression~\citep{lars} are common in feature
selection and can often result in performance guarantees, there are
only a similar NAS approaches~\citep{Liu2017ProgressiveNA}.
Such forward algorithms are attractive, when one wants to \emph{expand existing models} as extra computation becomes viable. Forward methods can utilize such extra computational resources without rebooting the training as in backward
methods and sparse optimization. Furthermore, forward methods
naturally result in a spectrum of models of various complexities to suitably choose from. Crucially, unlike backward approaches, forward methods do not need to specify a finite
search space up front making them more general and easier to use when  warm-starting from prior available models and for lifelong learning.

Specifically, inspired by early neural network growth work \citep{cascadecorr}, we propose a method (\Petridish) of growing networks from small to large, where we
opportunistically add shortcut connections in a fashion that is
analogous to applying gradient boosting~\citep{gboost} to the intermediate feature
layers. To select from the possible shortcut connections, we also
exploit sparsity-inducing regularization~\citep{lasso} during the training of the eligible shortcuts. 

We experiment with \Petridish for
both the cell-search~\citep{NASCell}, where we seek a shortcut
connection pattern and repeat it using a manually designed skeleton
network to form an architecture, and the less common but more general
macro-search, where shortcut connections can be freely formed.
Experimental results show \Petridish macro-search to be better than
previous macro-search NAS approaches on vision tasks, and brings
macro-search performance up to par with cell-search counter to beliefs
from other NAS works~\citep{nas,Pham2018EfficientNA} that macro-search
is inferior to cell-search.  \Petridish cell-search also finds models
that are more cost-efficient than those from~\citep{Liu2018DARTSDA},
while using similar training computation. This indicates that forward
selection methods for NAS are effective and useful.

We summarize our contribution as follows.
\begin{itemize}
\item We propose a forward neural architecture search algorithm that is analogous to gradient boosting on intermediate layers, allowing models to grow in complexity during training and warm-start from existing architectures and weights.
\item On CIFAR10 and PTB, the proposed method finds competitive models in few GPU-days with both cell-search and macro-search.
\item The ablation studies of the hyper-parameters highlight the importance
of starting conditions to algorithm performance.
\end{itemize}

\section{Background and Related Work}

\textbf{Sample-based.} \cite{nas} 
leveraged policy gradients~\citep{policygradient} to learn to 
sample networks, and established the now-common framework of
sampling networks and evaluating them after a few epochs of
training. The policy-gradient sampler 
has been replaced with evolutionary algorithms~\citep{schaffer1990using,Real2018RegularizedEF,Elsken2018EfficientMN},
Bayesian optimization~\citep{Kandasamy2018BNAS}, and Monte Carlo tree search~\citep{Negrinho2017DeepArchitectAD}. Multiple search-spaces~\citep{Elsken2018NeuralAS} are 
also studied under this framework. \cite{NASCell} introduce the idea of cell-search, where we learn a connection pattern, called a cell, and stack 
cells to form networks. \cite{Liu2018HierNA} further learn how to stack 
cells with hierarchical cells. \cite{CaiPathLevel} evolve networks 
starting from competitive existing models via net-morphism~\citep{netmorphism}.

\textbf{Weight-sharing.} The sample-based framework of~\citep{nas} spends most of its training computation 
in evaluating the sampled networks independently, and can
cost hundreds of GPU-days during search. 
This framework is revolutionized 
by ~\cite{Pham2018EfficientNA}, who share the weights of the possible networks and train all possible networks jointly. \cite{Liu2018DARTSDA} formalize NAS with weight-sharing as a bi-level optimization~\citep{bilevel_opt},
where the architecture and the model parameters are jointly learned. 
\cite{snas} leverage policy gradient to update the architecture in order to update the whole bi-level optimization with gradient descent. 

\textbf{Forward NAS.} Forward NAS originates from one of the earliest NAS works by \cite{cascadecorr} termed ``Cascade-Correlation'', in which, neurons are added to networks iteratively. Each new neuron takes input from existing neurons, and maximizes the correlation between its activation and the residual in network prediction. Then the new neuron is frozen and is used to improve the final prediction. 
This idea of iterative growth has been recently studied in~\citep{adanet,boostedresnet}
via gradient boosting~\citep{gboost}. While \Petridish is similar to gradient boosting, it is applicable to any layer, instead of only the final layer. 
Furthermore, \Petridish initializes weak learners without freezing or affecting the current model, unlike in gradient boosting, which freezes previous models.
\cite{Liu2017ProgressiveNA} have proposed forward search within the sampling
framework of ~\citep{nas}. \Petridish instead utilizes weight-sharing, reducing the search time from hundreds of GPU-days to just a few. 

\section{Preliminaries}
\label{sec:nas_bi_level_optimization}

\noindent \textbf{Gradient Boosting:}
Let $\mathcal{H}$ be a space of weak learners. 
Gradient boosting matches weak learners $h \in \mathcal{H}$ to 
the functional gradient $\nabla _{\hat{y}} \mathcal{L}$ of the loss $\mathcal{L}$ with respect to the prediction $\hat{y}$.
The weak learner $h^*$ that matches the negative gradient the best is added to the ensemble of learners, i.e.,
\begin{align}
h^* = \argmin _{h \in \mathcal{H}} \langle \nabla _{\hat{y}} \mathcal{L}, h \rangle.
\end{align}
Then the predictor is updated to become $\hat{y} \leftarrow \hat{y} + \eta h^*$, where $\eta$ is the learning rate.

\noindent\textbf{NAS Optimization:}
Given data sample $x$ with label $y$ from a distribution $\mathcal{D}$,  a neural network architecture $\alpha$ with parameters $w$ produces 
a prediction $\hat{y}(x ; \alpha, w)$ and suffers a prediction loss $\ell(\hat{y}(x ; \alpha, w), y)$.
The expected loss is then 
\begin{align}
\mathcal{L}(\alpha, w) = \mathbb{E} _{x, y \sim \mathcal{D}} [ \ell(\hat{y}(x ; \alpha, w), y) ] 
\approx \frac{1}{|\mathcal{D}_\textrm{train}|}
   \sum _{(x, y) \in \mathcal{D}_\textrm{train}} \ell(\hat{y}(x ; \alpha, w), y).
\end{align}
In practice,
the loss $\mathcal{L}$ is estimated on the empirical training data $\mathcal{D}_\textrm{train}$.
Following~\citep{Liu2018DARTSDA}, the problem of neural architecture search can be formulated as a bi-level optimization~\citep{bilevel_opt}
of the network architecture $\alpha$ and the model parameters $w$ under the loss $\mathcal{L}$ 
as follows.
\begin{align}
\min _{\alpha} \mathcal{L} (\alpha, w(\alpha)),
\quad
\textrm{s.t.} \quad w(\alpha) = \argmin _w \mathcal{L} (\alpha, w) 
\quad \textrm{and} \quad c(\alpha) \leq K,
\label{eq:bilevel_nas}
\end{align}
where $c(\alpha)$ is the test-time computational cost of the architecture, and $K$ is some constant. Formally, let $x_1, x_2, ...$ be intermediate layers in a feed-forward network. Then a shortcut from layer $x_i$ to $x_j$ ($j > i$) using operation $op$ is represented by $(x_i, x_j, op)$, where the operation $op$ is a unary operation such as 3x3 conv. We merge multiple shortcuts to the same $x_j$ with summation, unless specified otherwise using ablation studies. Hence, the architecture $\alpha$ is a collection of shortcut connections.

\textbf{Feature Selection Analogy:}
We note that Eq.~\ref{eq:bilevel_nas} generalizes feature selection for linear
prediction~\citep{lasso,omp,omp_fr}, where $\alpha$ selects feature subsets, $w$ is the prediction coefficient, and the loss is expected square error. 
Hence, we can understand a NAS algorithm by 
considering its application to feature selection, as discussed in the introduction and related works. This work draws a parallel to the feature selection algorithm Forward-Stagewise Linear Regression (FSLR)~\citep{lars} with small step sizes, which is an approximation to Least-angle Regression~\citep{lars}. 
In FSLR, we iteratively update with small step sizes the weight of the feature that correlates the most with the prediction residual. Viewing candidate features as weak learners, the residuals become the gradient of the square loss with respect to the linear prediction. Hence, FSLR is also understood as gradient boosting~\citep{gboost}.

\textbf{Cell-search vs. Macro-search:}
In this work, we consider both cell-search, 
a popular NAS search space where a network is a predefined sequence of some learned connection patterns~\citep{NASCell,Real2018RegularizedEF,Pham2018EfficientNA,Liu2018DARTSDA}, called cells,  
and macro-search, a more general NAS where no repeatable patterns are required. 
For a fair comparison between the two, we set both macro and cell searches to start with 
the same seed model, which consists of a sequence of simple cells. Both searches also choose from the same set of shortcuts.  The only difference is cell-search cells changing uniformly and macro-search cells changing independently.

\section{Methodology: Efficient Forward Architecture Search (\Petridish)}

Following gradient boosting strictly would limit the model growth to be only at the prediction layer of the network, $\hat{y}$. 
Instead, this work seeks to jointly expand the expressiveness of the network at intermediate layers, $x_1, x_2,...$. Specifically, we consider adding a weak learner $h_k \in \mathcal{H}_k$ at each $x_k$, where 
$\mathcal{H}_k$ (specified next) is the space of weak learners for layer $x_k$. 
$h_k$ helps reduce the gradient of the loss $\mathcal{L}$ with respect to $x_k$, $\nabla _{x_k} \mathcal{L}  = \mathbb{E} _{x, y \sim \mathcal{D}} [ \nabla _{x_{k}} \ell(\hat{y}(x ; \alpha, w), y) ]$, i.e., we choose $h_k$ with 
\begin{align}
h_k = \argmin _{h \in \mathcal{H}_k} \langle h, 
\nabla _{x_k} \mathcal{L} (\alpha, w) \rangle = 
\argmin _{h\in \mathcal{H}_k} \langle h, \mathbb{E} _{x, y \sim \mathcal{D}} [ \nabla _{x_{k}} \ell(\hat{y}(x ; \alpha, w), y) ] \rangle.
\label{eq:hallu_objective}
\end{align}
Then we expand the model by adding $h_k$ to $x_k$.  In other words, we replace each $x_k$ with $x_k + \eta h_k$ in the original network, where $\eta$ is a scalar variable initialized to 0. 
The modified model then can be trained with backpropagation. We next specify the weak learner space, and how they are learned.

\textbf{Weak Learner Space:} The weak learner space $\mathcal{H}_{k}$ for a layer $x_{k}$ is formally 
\begin{align}
\mathcal{H}_{k} = \{ \texttt{op}_{\textrm{merge}}( \texttt{op}_1(z_1), ..., \texttt{op}_{I_{\textrm{max}}}(z_{I_{\textrm{max}}})) : z_1, ..., z_{I_{\textrm{max}}} \in \texttt{In}(x_{k}), \texttt{op}_1, ..., \texttt{op}_{I_{\textrm{max}}} \in \texttt{Op}  \},
\label{eq:hallu_space}
\end{align}
where $\texttt{Op}$ is the set of eligible unary operations, 
$\texttt{In}(x_{k})$ is the set of allowed input layers, $I_{\textrm{max}}$ is the
number of shortcuts to merge together in a weak learner, and $\texttt{op}_{\textrm{merge}}$ is 
a merge operation to combine the shortcuts into a tensor of the same shape as $x_k$. 
On vision tasks, following~\citep{Liu2018DARTSDA}, we set $\texttt{Op}$ to contain
separable conv \texttt{3x3} and \texttt{5x5}, dilated conv \texttt{3x3} and \texttt{5x5}, max and average pooling \texttt{3x3}, and identity. The separable conv is applied twice as per~\citep{Liu2018DARTSDA}. Following~\citep{NASCell,Liu2018DARTSDA}, we set $\texttt{In}(x_{k})$ to be layers that are topologically earlier than $x_k$, and are either in the same cell as $x_k$ or the outputs of the previous two cells. 
We choose $I_{\textrm{max}}=3$ through an ablation study 
from amongst 2, 3 or 4 in Sec.~\ref{sec:experiment_number_operations}, and 
we set $\texttt{op}_{\textrm{merge}}$ to be a concatenation followed by a projection with conv \texttt{1x1} through an ablation study in Sec.~\ref{sec:sum_vs_cat_proj} against weighted sum. 

\begin{algorithm}[t]
\begin{algorithmic}[1]
\STATE \textbf{Input}: 
(1) $L_x$, the list of layers in the current model (macro-search) or current cell (cell-search) in topological order;
(2) $\texttt{is\_out}(x)$, whether we are to expand at $x$;
(3) $\lambda$, hyper parameter for selection shortcut connections. 
\STATE \textbf{Output}: (1) $L'_x$, the modified $L_x$ with weak learners $x_c$; 
(2) $L_c$, the list of $x_c$ created;
(3) $\ell_{extra}$, the additional training loss.

\STATE $L'_x \leftarrow L_x$; \;\; $L_c \leftarrow \text{empty list}$; \;\; $\ell_{extra} \leftarrow 0$ 
\FOR{$x_k$ in enumerate($L_x$)}
    \STATE  \textbf{if} { not \texttt{is\_out}($x_{k}$)} \; \textbf{then} \; continue  \; \textbf{end if}
    \STATE Compute the eligible inputs $\text{In}(x_{k})$, and index them as $z_1,...,z_I$.

    \STATE $x_c \leftarrow \sum _{i=1}^I \sum _{j=1}^J  \alpha^k_{i,j}\texttt{op}_j(\stopgradient (z_i))$.
    \label{algline:add_sg}
\STATE Insert the layer $x_c$ right before $x_{k}$ in $L'_x$.
\STATE $\ell_{extra} \leftarrow \ell_{extra} + \lambda \sum _{i=1}^I \sum _{j=1}^J |\alpha^k_{i,j}|$.
\STATE Append $x_c$ to $L_c$.
\STATE Modify $x_{k}$ in $L'_x$ so that $x_{k} \leftarrow x_{k} + \stopforward (x_c)$.
\label{algline:add_sf}
\ENDFOR
\end{algorithmic}
\caption{\Petridish .initialize\_candidates}
\label{alg:candidate_init}
\end{algorithm}

\begin{figure}[ht]
\centering
\subfloat[]{
    \includegraphics[height=0.34\textwidth, keepaspectratio]{\NASDIR/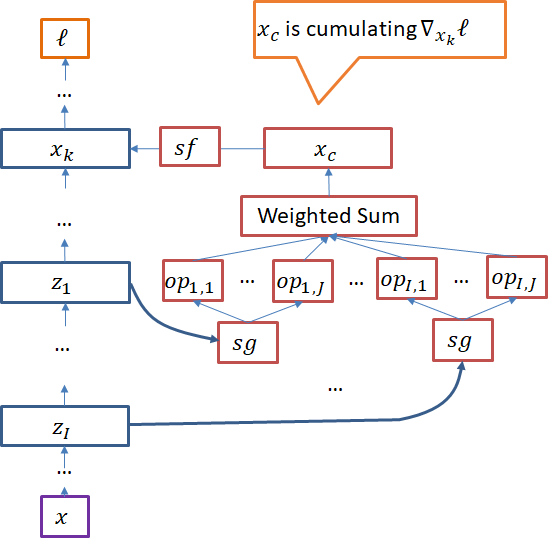}
    \label{fig:x_c_select_sf_sg}
}
~
\subfloat[]{
\includegraphics[height=0.34\textwidth, keepaspectratio]{\NASDIR/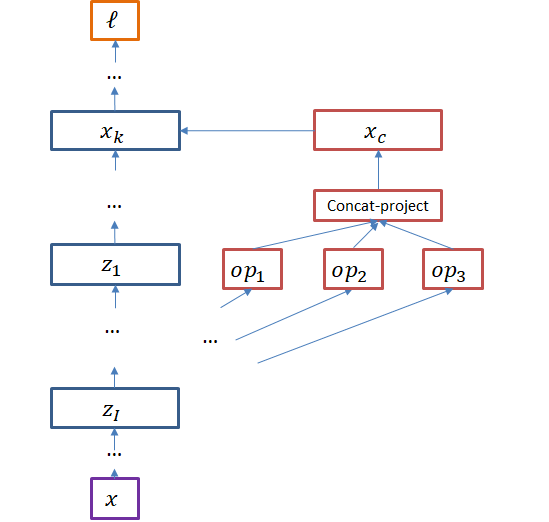}
\label{fig:x_c_select_final}
}
    \caption{(a) Blue boxes are in the parent model, and red boxes are for weak learning. 
    Operations are joined together in a weighted sum to form $x_c$, in order to match 
    $-\nabla _{x_k} \mathcal{L}$. 
    (b) The top $I_{\textrm{max}}$ operations are selected and merged with a concatenation, followed by a projection. 
    }
\end{figure}

\textbf{Weak Learning with Weight Sharing:}
\label{sec:candidate_init_and_select}
In gradient boosting, one typically optimizes Eq.~\ref{eq:hallu_objective} by minimizing
$\langle h, \nabla_{x_k} \mathcal{L} \rangle$ for multiple $h$, and selecting the best $h$ afterwards.
However, as there are $\binom{IJ}{I_{\textrm{max}}}$ possible weak learners in the space of Eq.~\ref{eq:hallu_space}, where $I = |\texttt{In}(x_{k})|$ and $J = |\texttt{Op}|$,
it may be costly to enumerate all possibilities.
Inspired by the parameter sharing works in NAS~\citep{Pham2018EfficientNA,Liu2018DARTSDA} and model compression in neural networks~\citep{huang2017condensenet},
we propose to jointly train the union of all weak learners, while learning to select the shortcut connections. This process also only costs a constant factor more than training one weak learner. Specifically, we fit the following joint weak learner $x_c$ for a layer $x_k$ in order to minimize 
$\langle x_c, \nabla _{x_k} \mathcal{L} \rangle$: 
\begin{align}
    x_c = \sum _{i=1}^I \sum_{j=1}^J \alpha_{i,j} \texttt{op}_j(z_i),
    \label{eq:x_c_select}
\end{align}
where $\texttt{op}_j \in \texttt{Op}$ and $z_i \in \texttt{In}(x_k)$ enumerate all possible operations and inputs, and $\alpha_{i,j} \in \mathbb{R}$ is the weight of the shortcut $\texttt{op}_j(z_i)$. 
Each $\texttt{op}_j(z_i)$ is normalized with batch-normalization to have approximately zero mean and unit variance in expectation, so $\alpha_{i,j}$ reflects 
the importance of the operation.
To select the most important operations, we minimize 
$\langle x_c, \nabla _{x_k} \mathcal{L} \rangle$ with an $L1$-regularization on the weight vector $\vec{\alpha}$, i.e., 
\begin{align}
    \lambda \Vert \vec{\alpha} \Vert_1 = 
    \lambda \sum_{i=1}^I \sum _{j=1}^J | \alpha _{i,j} |,
    \label{eq:x_c_select_loss}
\end{align} 
where $\lambda$ is a hyper-parameter which we choose in the appendix~\ref{sec:l1_lambda_select}. $L1$-regularization, known as Lasso~\citep{lasso}, induces sparsity in the parameter and is widely used for feature selection.

\textbf{Weak Learning Implementation:} 
A na\"ive implementation of joint weak learning needs to compute $\nabla_{x_k}\mathcal{L}$ and freeze the existing model during weak learner training. Here we provide a modification to avoid these two costly requirements. 
Algorithm~\ref{alg:candidate_init} describes the proposed implementation 
and Fig.~\ref{fig:x_c_select_sf_sg} illustrates the weak learning computation graph. 
We leverage a custom operation called stop-gradient, \stopgradient\xspace, which has the property that for any $x$, $\stopgradient(x) = x$ and $\nabla _x \stopgradient(x) = 0$. Similarly, we define the complimentary operation stop-forward, $\stopforward(x) = x - \stopgradient(x)$, i.e., $\stopforward(x) = 0$ and $\nabla _x \stopforward(x) = \textrm{Id}$, the identity function.
Specifically, on line~\ref{algline:add_sg}, we apply $\stopgradient$ to inputs of weak learners, so that $x_c = \sum _{i=1}^I \sum_{j=1}^J \alpha_{i,j} \texttt{op}_j(\stopgradient(z_i))$ does not affect the gradient of the existing model. 
Next, on line~\ref{algline:add_sf}, we replace the layer $x_k$ with $x_k +
\stopforward(x_c)$, so that the prediction of the model is unaffected by weak learning. 
Finally, the gradient of the loss with respect to any weak learner parameter $\theta$ is:
\begin{align}
\nabla_{\theta} \mathcal{L} &= \nabla _{x_k + \stopforward(x_c)} \mathcal{L} \nabla _{x_c} \stopforward(x_c) \nabla _{\theta} x_c = \nabla _{x_k}\mathcal{L} \nabla _{\theta} x_c = \nabla_{\theta} \langle \nabla_{x_{k}} \mathcal{L}, x_c \rangle.
\end{align}
This means that $\stopforward$ and $\stopgradient$ not only prevent the weak learning from affecting the training of existing model, but also enable us to minimize
$\langle \nabla_{x_{k}} \mathcal{L}, x_c\rangle$ via backpropagation on the whole network. Thus, we no longer need explicitly compute $\nabla_{x_{k}} \mathcal{L}$ nor freeze the existing model weights during weak learning. Furthermore, since weak learners of different layers do not interact during weak learning, we grow the network at all $x_k$ that are ends of cells at the same time.

\begin{algorithm}[t]
\begin{algorithmic}[1]
\STATE \textbf{Inputs}: (1) $L'_x$, the list of layers of the model in topological order;
(2) $L_c$, list of selection modules in $L'_x$;
(3) $\alpha^k_{i,j}$, the learned operation weights of $x_c$ for layer $x_k$.
\STATE \textbf{Output}: A modified $L'_x$, which is to be trained with backpropagation for a few epochs.
\FOR{$x_c$ in $L_c$}
    \STATE Let $A = \{\alpha^{k}_{i,j}: i = 1,..., I, j = 1,..., J\}$  be the weights of operations in $x_c$.
    \STATE Sort $\{ |\alpha| : \alpha \in A \}$, and let $\texttt{op}_1, ..., \texttt{op}_{I_{\textrm{max}}}$ be operations with the largest associated $|\alpha|$.
    \STATE Replace $x_c$ with $\texttt{proj}(\texttt{concat}(\texttt{op}_1, ..., \texttt{op}_{I_{\textrm{max}}}))$ in $L'_x$. $\texttt{proj}$ is to the same shape as $x_k$.
\ENDFOR
\STATE Remove all $\stopgradient (\cdot)$. Replace each $\stopforward (x)$ with a $\eta x$, where 
$\eta$ is a scalar variable initialized to 0.
\end{algorithmic}
\caption{\Petridish .finalize\_candidates}
\label{alg:candidate_select}
\end{algorithm}

\textbf{Finalize Weak Learners:}
\label{sec:candidate_finalize}
In Algorithm~\ref{alg:candidate_select} and Fig.~\ref{fig:x_c_select_final}, we finalize the weak learners. We select in each $x_c$ the top $I_{\textrm{max}}$ shortcuts according to the absolute value of $\alpha_{i,j}$, and merge them with a concatenation followed by a projection to the shape of $x_k$. We note that the weighted sum during weak learning 
is a special case of concatenation-projection, and we use an ablation study in appendix~\ref{sec:sum_vs_cat_proj} to validate this replacement. We also note that most NAS works ~\citep{NASCell,Real2018RegularizedEF,Pham2018EfficientNA,Liu2018DARTSDA,snas,proxyless} have similar set-ups of concatenating intermediate layers in cells and projecting the results.
We train the finalized models for a few epochs, warm-starting from the parameters in weak learning.

\textbf{Remarks:} 
A key design concept of \Petridish is amortization, where we require the computational costs of weak learning and model training to be a constant factor of each other. We further
design \Petridish to do both at the same time. 
Following these principles, it only costs a constant factor of additional computation
to augment models with \Petridish while training the model concurrently.

We also note that since \Petridish only grows models, noise in weak learning and model training can result in sub-optimal short-cut selections. To mitigate this potential problem and to reduce the search variance, we utilize multiple parallel workers of \Petridish, each of which can warm-start from intermediate models of each other. We defer this implementation detail to the appendix.

\section{Experiments}
\label{sec:nas_experiment}


We report the search results on CIFAR-10~\citep{cifar} and the transfer result on ImageNet~\citep{ILSVRC15}. Ablation studies for choosing the hyper parameters are deferred to appendix~\ref{sec:ablation_studies},
which also demonstrates the importance of blocking the influence of weak learners to the existing models during weak learning via $\stopforward$ and $\stopgradient$. We also search on Penn Tree Bank~\citep{ptb}, and show that it is not an interesting data-set for evaluating NAS algorithms.

\vspace{-2pt}
\subsection{Search Results on CIFAR10}
\label{sec:experiment_cifar10_search}
\vspace{-2pt}

\noindent \textbf{Set-up:} Following~\citep{NASCell,Liu2018DARTSDA}, we search on a shallow and slim  networks, which have $N=3$ normal cells in each of the three feature map resolution, one transition cell  between each pair of adjacent resolutions, and $F=16$ initial filter size. Then we scale up the found model to have $N=6$ and $F=32$ for a final training from scratch.
During search, we use the last 5000 training images as a validation set. 
The starting seed model is a modified ResNet~\citep{resnet}, where the output of a cell is the sum of the input and the result of applying two \texttt{3x3} separable conv to the input. This is one of the simplest seeds in the search space popularized by~\citep{NASCell,Pham2018EfficientNA,Liu2018DARTSDA}.
The seed model is trained for 200 epochs, with a batch size of 32 and a learning rate that decays from 0.025 to 0 in cosine decay~\citep{cosine_lr}. We apply drop-path~\citep{larsson2016fractalnet} with probability 0.6 and the standard CIFAR-10 cut-out~\citep{cutout}. Weak learner selection and finalization are trained for 80 epochs each, using the same parameters. The final model training is from scratch for 600 epochs on all training images with the same parameters.

\begin{table*}[t]
    \centering
    \caption{Comparison against state-of-the-art recognition results on CIFAR-10. Results marked with $\dagger$ are not trained with cutout. The first block represents approaches for macro-search. The second block represents approaches for cell-search. 
    We report \Petridish results in the format of ``best $|$ mean $\pm$ standard deviation'' among five repetitions of the final training.
    }
    \begin{tabular}{l|cccc}
    \hline
\multirow{ 2}{*}{\textbf{Method} }
        &  \textbf{\# params} 
        &  \textbf{Search } 
        &  \textbf{Test Error } \\
        &  (mil.)
        &  (GPU-Days)
        &  (\%)\\
\hline
\citet{nas}$^{\dagger}$
    &  7.1 &  1680+ &  4.47  \\
\citet{nas} + more filters$^{\dagger}$
    &  37.4 &   1680+ &  3.65   \\
\citet{Real2017EvoNet}$^{\dagger}$
    &  5.4 &   2500 &  5.4  \\
ENAS macro~\citep{Pham2018EfficientNA}$^{\dagger}$
    &  21.3 &  0.32 &  4.23 \\
ENAS macro + more filters$^{\dagger}$
    &  38 &   0.32 &  3.87 \\
Lemonade I~\citep{Elsken2018EfficientMN}
    &  8.9 &    56 &  3.37 \\
\hline
\Petridish initial model ($N=6$, $F=32$)
    & 0.4 &  -- & 4.6 \\
\textbf{\Petridish macro} 
    & 2.2 & 5 & 2.83 $|$ 2.85$\pm$ 0.12 \\
\hline \hline
NasNet-A~\citep{NASCell}
    &  3.3 &    1800 &  2.65   \\
AmoebaNet-A~\citep{Real2018RegularizedEF}
    &  3.2 &  3150 &  3.3  \\
AmoebaNet-B~\citep{Real2018RegularizedEF} 
    &  2.8 &   3150 &  2.55 \\ 
PNAS~\citep{Liu2017ProgressiveNA}$^{\dagger}$
    &  3.2 &  225 &  3.41 \\
ENAS cell~\citep{Pham2018EfficientNA}
    &  4.6 &  0.45 &  2.89 \\ 
Lemonade II~\citep{Elsken2018EfficientMN}
    &  3.98 &  56 &  3.50 \\
Darts~\citep{Liu2018DARTSDA}
    &  3.4 &   4 &  2.83 \\ 
Darts random~\citep{Liu2018DARTSDA}
    & 3.1 & -- & 3.49 \\
\citet{NAONet}$^{\dagger}$
    & 3.3 & 0.4 & 3.53 \\
PARSEC \citep{parsec}
    & 3.7  & 1 & 2.81 \\
\hline
\textbf{\Petridish cell}
    & 2.5 & 5 & 2.61 $|$ 2.87 $\pm$ 0.13 \\
\textbf{\Petridish cell more filters (F=37)}
    & 3.2 & 5 &  2.51 $|$  2.75 $\pm$ 0.21 \\
\hline
    \end{tabular}
    \label{tab:cifar10_search}
\end{table*}

\begin{table*}[t]
    \centering
    \caption{The performance of the best CIFAR model transferred to ILSVRC. 
    Variance is from multiple training of the same model from scratch. 
    }
    \label{tab:imagenet_compare}
    \resizebox{\textwidth}{!}{
    \begin{tabular}{l|cccc}
    \hline
\multirow{2}{*}{\textbf{Method} }
        &  \textbf{\# params} 
        &  \textbf{\# multi-add}
        &  \textbf{Search}
        &  \textbf{top-1 Test Error } \\
        &  (mil.)
        &  (mil.)
        &  (GPU-Days)
        &  (\%)\\
\hline
Inception-v1 (Szegedy et al., 2015)
    & 6.6 & 1448 & -- & 30.2 \\
MobileNetV2 (Sandler et al., 2018)
    & 6.9 & 585 & -- & 28.0 \\
\hline
NASNet-A (Zoph et al., 2017) 
    & 5.3 & 564 & 1800 & 26.0 \\
NASNet-B (Zoph et al., 2017) 
    & 5.3 & 488 & 1800 & 27.2 \\
AmoebaNet-A (Real et al., 2018)
    & 5.1 & 555 & 3150 & 25.5 \\
Path-level (Cai et al., 2018)
    & -- & 588 & 8.3 & 25.5 \\
PNAS (Liu et al., 2017a)
    & 5.1 & 588 & 225  & 25.8 \\
DARTS (Liu et al., 2019)
    & 4.9 & 595 & 4    & 26.9 \\
SNAS \citep{snas}
    & 4.3 & 522 & 1.6 & 27.3 \\
Proxyless \citep{proxyless}
    & 7.1 & 465 & 8.3 & 24.9 \\
PARSEC \citep{parsec}
    & 5.6 & -- & 1  & 26.0 \\
\hline
\textbf{\Petridish macro} (N=6,F=44) 
    & 4.3 & 511 & 5 & 28.5 $|$ 28.7 $\pm$ 0.15\\
\hline
\textbf{\Petridish cell} (N=6,F=44) 
    & 4.8 & 598 & 5 & 26.0 $|$ 26.3 $\pm$ 0.20 \\
\hline
\end{tabular}
} 
\end{table*}
\noindent \textbf{Search Results:} Table~\ref{tab:cifar10_search} depicts the test-errors, model parameters, and search computation of the proposed methods along with many state-of-the-art methods.
We mainly compare against models of fewer than 3.5M parameters, since these models can be easily transferred to ILSVRC~\citep{ILSVRC15} mobile setting via a standard procedure~\citep{NASCell}.
The final training of \Petridish models is repeated five times. 
\Petridish cell search finds a model with 2.87$\pm$0.13\% error rate with 2.5M parameters, in 5 GPU-days using GTX 1080. Increasing filters to $F=37$, the model has 2.75$\pm$0.21\% error rate with 3.2M parameters.
This is one of the better models among models that have fewer than 3.5M parameters, and is in particular better than DARTS~\citep{Liu2018DARTSDA}. 

\Petridish macro search finds a model that achieves 2.85$\pm$ 0.12\% error rate using 2.2M parameters in the same search computation. This is significantly better than previous macro search results, and showcases that macro search can find cost-effective architectures that are previously only found through cell search. 
This is important, 
because the NAS literature has been moving away from macro architecture search, as early
works~\citep{NASCell,Pham2018EfficientNA,Real2018RegularizedEF} have shown that 
cell search results tend to be superior to those from macro search.
However, this result may be explained by the superior initial models of cell search: the initial model of \Petridish is one of the simplest models that any of the listed cell search methods proposes and evaluates, and it already achieves 4.6\% error rate using only 0.4M parameters, a result already on-par or better than any other macro search result. 

We also run multiple instances of \Petridish cell-search to study the variance in search results, and  Table~\ref{tab:multiple_search_results} reports performance of the best model of each search run. We observe that the models from the separate runs have similar performances. Averaging over the runs, the search time is 10.5 GPU-days and the model takes 2.8M parameters to achieve 2.88\% average mean error rate. Their differences may be caused by the randomness in stochastic batches, variable initialization, image pre-processing, and drop-path. 

\begin{table}[t]
  \begin{minipage}[]{0.49\textwidth}
    \centering
    \captionof{table}{Performances of the best models from multiple instances of \Petridish cell-search.}
    \label{tab:multiple_search_results}
    \begin{tabular}{cccc}
    \hline
          \textbf{\# params} 
        &  \textbf{Search}
        &  \textbf{Test Error } \\
          (mil.)
        &  (GPU-Days)
        &  (\%)\\
    \hline
        3.32 & 7.5 & 2.80 $\pm$ 0.10 \\
        2.5 & 5 & 2.87 $\pm$ 0.13 \\
        2.2 & 12 & 2.88 $\pm$ 0.15 \\
        2.61 & 18 & 2.90 $\pm$ 0.12 \\
        3.38 & 10 & 2.95 $\pm$ 0.09 \\
    \hline
    \end{tabular}
  \end{minipage}
  \hfill
  \begin{minipage}[]{0.49\textwidth}
    \centering
    \includegraphics[width=\linewidth,keepaspectratio]{\NASDIR/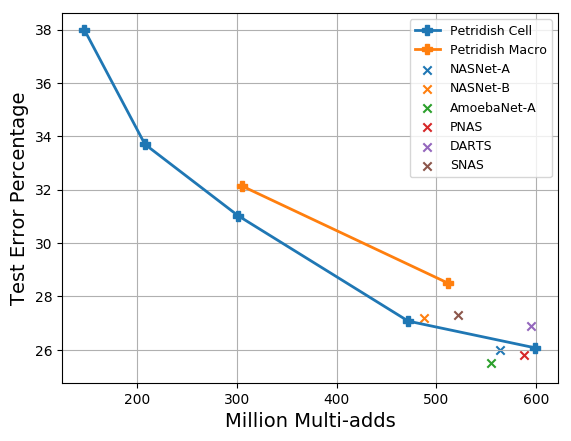}
    \captionof{figure}{\Petridish naturally find a collection of models of different complexity
    and accuracy. Models outside of the lower convex hull are removed for clarity.}%
  \label{fig:imagenet_convexhull}
  \end{minipage}
\end{table}

\noindent \textbf{Transfer to ImageNet:} 
We focus on the mobile setting for the model transfer results on ILSVRC~\citep{ILSVRC15}, which means we limit the number of multi-add per image to be within 600M. 
We transfer the final models on CIFAR-10 to ILSVRC by adding an initial \texttt{3x3} conv of stride of 2, followed by two transition cells, to down-sample the \texttt{224x224} input images to \texttt{28x28} with $F$ filters. In macro-search, where no transition cells are specifically learned, we again use the the modified ResNet cells from the initial seed model as the replacement. After this initial down-sampling, the architecture is the same as in CIFAR-10 final models. Following~\citep{Liu2018DARTSDA}, we train these models for 250 epochs with batch size 128, weight decay $3*10^{-5}$, and initial SGD learning rate of 0.1 (decayed by a factor of 0.97 per epoch). 

Table~\ref{tab:imagenet_compare} depicts performance of the transferred models. The \Petridish cell-search model achieves 26.3$\pm$0.2\% error rate using 4.8M parameters and 598M multiply-adds, which is on par with state-of-the-art results listed in the second block of Table~\ref{tab:imagenet_compare}. By utilizing feature selection techniques to evaluate multiple model expansions at the same time, \Petridish is able to find models faster by one or two orders of magnitude than early methods that train models independently, such as NASNet~\citep{NASCell}, AmoebaNet~\citep{Real2018RegularizedEF}, and PNAS~\citep{Liu2017ProgressiveNA}.  
In comparison to super-graph methods such as DARTS~\citep{Liu2018DARTSDA}, \Petridish cell-search takes similar search time to find a more accurate model.

The \Petridish macro-search model achieves 28.7$\pm$0.15\% error rate using 4.3M parameters and 511M multiply-adds, a comparable result to the human-designed models in the first block of Table~\ref{tab:imagenet_compare}. Though this is one of the first successful transfers of macro-search result on CIFAR to ImageNet, the relative performance gap between cell-search and macro-search widens after the transfer. This may be because the default transition cell is not adequate for transfer to more complex data-sets. 
As \Petridish gradually expands existing models, we naturally receive a gallery of models of various computational costs and accuracy. Figure~\ref{fig:imagenet_convexhull} showcases the found models.

\vspace{-2pt}
\subsection{Search Results on Penn Treebank}
\vspace{-2pt}

\Petridish when used to grow the cell of a recurrent neural network achieves a best test perplexity of $55.85$ and average test perplexity of $56.39\pm0.38$ across $8$ search runs with different random seeds on PTB. This is competitive with the best search result of \citep{randnas} of $55.5$ via random search with weight sharing. In spite of good performance we don't put much significance on this particular language-modeling task with this data set because no NAS algorithm appears to perform better than random search \citep{randnas}, as detailed in appendix~\ref{sec:ptb_results}.

\vspace{-2pt}
\section{Conclusion}
\label{sec:discussion}
\vspace{-2pt}
We formulate NAS as a bi-level optimization problem, which generalizes feature 
selection for linear regression. We propose an efficient forward selection algorithm that applies gradient boosting to intermediate layers, and generalizes the feature selection algorithm LARS~\citep{lars}. We also speed weak learning via weight sharing, training the union of weak learners and selecting a subet from the union via $L1$-regularization. We demonstrate experimentally that forward model growth can find accurate models in a few GPU-days via cell and macro searches.
Source code for \Petridish is available at \url{https://github.com/microsoft/petridishnn}.

\bibliographystyle{icml.bst}
\bibliography{network_search}

\newpage
\appendix

\section{Additional Implementation Details}
\subsection{Parallel Workers} 
\label{sec:parallel_workers}
Since there are many sources of randomness in model training and weak learning, including SGD batches, drop-path, cut-out, and variable initialization, \Petridish can benefit from multiple runs. Furthermore, if one worker finds a cost-efficient model of a medium size, other workers may want the option to warm-start from this checkpoint. \Petridish workers warm-start from models on the lower convex hull of the scatter plot of model validation error versus model complexity, because any mixture of other models are either more complex or less accurate. 

As there are multiple models on the convex hull, the workers need also choose one at each iteration. To do so, we loop over the models on the hull from the most accurate to the least, 
and choose a model $m$ with a probability $\frac{1}{n(m)+1}$, where $n(m)$ is the number of times that $m$ is already chosen. This probability is chosen because if a model has been sampled $n$ times, then the next child is the best among the $n+1$ children with probability $\frac{1}{n+1}$. We favor the accurate models, because it is typically more difficult to improve accurate models. In practice, \Petridish sample fewer than 100 models, so performances of different sampling algorithms are often indistinguishable, and we settle on this simple algorithm.

\subsection{Select Models for Final Training}
The search can be interrupted at anytime, and the best models are the models on the performance
convex hull at the time of interruption. For evaluating \Petridish on CIFAR-10~\citep{cifar}, we perform final training on models that are on the search-time convex hull and have near 60 million multi-adds on CIFAR-10 during search with $N=3$ and $F=16$. We focus on these models can be translated to the ILSVRC mobile setting easily with a fixed procedure of setting $N=6$ and $F=44$.

\subsection{Computation Resources}
The search are performed on docker containers that have access to four GPUs. The final training of CIFAR~\citep{cifar} and PTB~\citep{ptb} models each uses one GPUs. The final training of transferred models on ILSVRC each uses four GPUs. The GPUs can be V100, P100, or GTX1080. 

\section{Ablation Studies}
\label{sec:ablation_studies}

\subsection{Evaluation Criteria}
On CIFAR-10~\citep{cifar}, we often find that standard deviation of final training and 
search results to be high in comparison to the difference among different search 
algorithms. In contrast, the test-error on ILSVRC is more stable, and so that 
one can more clearly differentiate the performances of models from different search algorithms.
Hence, we use ILSVRC transfer results to compare search algorithms whenever the results are available. 
We use CIFAR-10 final training results to compare search algorithms, if otherwise. 

\subsection{Search Space: Direct versus Proxy}
\label{sec:direct_vs_proxy}
This section provides an ablation study on a common theme of recent 
neural architecture search works, where the search is conducted on a proxy space of small and shallow models, with results transferred to 
larger models later. In particular, since \Petridish uses iterative growth, it need not consider the complexity of a super graph containing all possible models. Thus, \Petridish can be applied directly to the final model setting on CIFAR-10, where $N=6$ and $F=32$. However, this implies each model takes about eight times the computation, and may introduce extra difficulty in convergence. Table~\ref{tab:direct_vs_proxy} shows the transfer results of the two approaches to ILSVRC.  We see that
this popular proxy search heuristic indeed leads to more accurate models. 

\begin{table}[t]
    \centering
    \begin{tabular}{l|cccc}
    \hline
\multirow{ 2}{*}{\textbf{Method} }
        &  \textbf{\# params} 
        &  \textbf{\# multi-add}
        &  \textbf{Search}
        &  \textbf{top-1 Test Error } \\
        &  (mil.)
        &  (mil.)
        &  (GPU-Days)
        &  (\%)\\
    \hline
    \Petridish cell direct (F=40) 
        & 4.4 & 583 & 15.3 &  26.9 \\
    \textbf{\Petridish cell proxy (F=44)} 
        & 4.8 & 598 & 5 & 26.3 \\
    \hline
    \end{tabular}
    \caption{Search space comparison between the direct space of $N=6$ and $F=32$ and the proxy space of $N=3$ and $F=16$ by evaluating their best mobile setting models on ILSVRC.}
    \label{tab:direct_vs_proxy}
\end{table}

\subsection{$\texttt{op}_{\textrm{merge}}$: Weighted Sum versus Concatenation-Projection}
\label{sec:sum_vs_cat_proj}

\begin{table}[t]
    \centering
    \caption{ILSVRC2012 transfer results. 
    	Ablation study on the choice of weighted-sum (WS), concat-projection at the end (CP-end), or the \Petridish default merge operation in finalized weak learners.
    	The searches were directly on the search space where $N=6$ and $F=32$.}
    \begin{tabular}{l|cccc}
    \hline
\multirow{ 2}{*}{\textbf{Method} }
        &  \textbf{\# params} 
        &  \textbf{\# multi-add}
        &  \textbf{Search}
        &  \textbf{top-1 Test Error } \\
        &  (mil.)
        &  (mil.)
        &  (GPU-Days)
        &  (\%)\\
\hline
WS macro(F=48) 
    & 5.9 & 756 & 29.5 & 32.5\\
CP-end macro (F=36) 
    & 5.4 & 680 & 29.5 & 29.1 \\
{\Petridish macro} (F=32) 
    & 4.9 & 593 & 27.2 & 29.4 \\
\hline
WS cell (F=48) 
    & 3.3 & 477 & 22.8 & 32.7\\
CP-end cell  (F=44) 
    & 4.7 & 630 & 22.8 & 27.2 \\
\textbf{\Petridish cell} (F=40) 
    & 4.4 & 583 & 15.3 &  26.9 \\
\hline  
\end{tabular}
\label{tab:imagenet_ws_vs_cp}
\end{table}

After selecting the shortcuts in Sec.~\ref{sec:candidate_finalize}, we concatenate them and project the result with 1x1 conv so that the result can be added to the output layer $x_{out}$. 
Here we empirically justify this design choice through consideration of two alternatives.  
We first consider applying the switch only to the final reported model.  In other words, instead of using concatenation-projection as the merge operation during search we switch all weak learner weighted-sums to concatenation-projections in the final model, which are trained from scratch to report results.  We call this variant CP-end.  Another variant where we never switch to concatenation-projection is called WS. Since concatenation-projection incurs additional computation to the model, we increase the channel size of WS variants so that the two variants have similar test-time multiply-adds for fair comparisons. The default \Petridish option is switching the weak learner weighted-sums to concatenation-projections each time weak learners are finalized with Alg.~\ref{alg:candidate_select}.  We compare WS, CP-end and \Petridish on the transfer results on ILSVRC in Table~\ref{tab:imagenet_ws_vs_cp}, and observe that \Petridish achieves similar or better prediction error using less test-time computation and training-time search.

\subsection{Is Weak Learning Necessary?}
\label{sec:experiment_soft_vs_hard}

\begin{table}[t]
    \centering
    \caption{ILSVRC2012 transfer results. 
    	Ablation study on the choice of \petridishsoft and \petridishhard for training the weak learners. 
    	The search were directly on the search space of $N=6$ and $F=32$, different from the proxy
    	space ($N=3, F=16$) used in the main text.
    }
    \begin{tabular}{l|cccc}
    \hline
\multirow{ 2}{*}{\textbf{Method} }
        &  \textbf{\# params} 
        &  \textbf{\# multi-add}
        &  \textbf{Search}
        &  \textbf{top-1 Test Error } \\
        &  (mil.)
        &  (mil.)
        &  (GPU-Days)
        &  (\%)\\
\hline
\Petridish \petridishsoft cell (F=32) 
    & 4.0 & 546 & 20.6 & 32.8 \\
\textbf{\Petridish cell} (F=40) 
    & 4.4 & 583 & 15.3 &  26.9 \\
\hline
\end{tabular}
\label{tab:imagenet_soft_vs_hard}
\end{table}

An interesting consideration is whether to stop the influence of the
weak learners to the models during the weak learning. On the one hand, we
eventually want to add the weak learners into the model and allow them
to be backpropagated together to improve the model accuracy. On the
other hand, the introduction of untrained weak learners to trained
models may negatively affect training. Furthermore, the models may
develop dependency on weak-learner shortcuts that are not selected,
which can also negatively affect future models. To study the
effects through an ablation study, we remove $\stopgradient$
and replace $\stopgradient$ with a variable scalar multiplication 
that is initialized to zero in
Algorithm~\ref{alg:candidate_init}. 
This is equivalent to adding the joint weak learner $x_c$ of 
Eq.~\ref{eq:x_c_select} directly to the boosted layer $x_k$ after 
random initialization, and then we train the existing model and the joint
weak learner together with backpropagation.
We call this variant \petridishsoft, and
compare it against the default \Petridish.
Table~\ref{tab:imagenet_soft_vs_hard} showcases the transfer results
of \petridishhard and \petridishsoft to ILSVRC. We compare
\Petridish cell (F=40) with \petridishsoft cell (F=32), two models
that have similar computational cost but very different accuracy, and
we observe that \petridishhard leads to much better model than
\petridishsoft for cell-search. This suggests that the 
randomly initialized joint weak learners should not directly be
added to the existing model to be backpropagated, and the weak learning
step is beneficial for the overall search.

\subsection{Number of Merged Operations, $I_{\textrm{max}}$}
\label{sec:experiment_number_operations}

\begin{table}[t]
    \centering
    \caption{Test error rates on CIFAR-10 by models found with different weak learner complexities.
    }
    \begin{tabular}{c|c}
    \hline
    $I_{\textrm{max}}$ & Average Lowest Error Rate \\
    \hline
    2 & 3.08  \\ 
    \textbf{3} & 2.88 \\  
    4 & 2.93 \\ 
    \hline
    \end{tabular}
    \label{tab:cifar10_search_choose_i}
\end{table}

As we initialize all possible shortcuts during weak learning, we need decide $I$, the number of them to select for forming the weak learner. On one hand, adding complex weak learners can boost performance rapidly. On the other, this may add sub-optimal weak learners that hinder future growth. We test the choice of $I=2,3,4$ during search. We run with each choice five times, and take the average of their most accurate models that take under 60 million multi-adds on the CIFAR model with $N=3$ and $F=16$. Models in this range are chosen, because their transferred models to ILSVRC can have 600 million multi-adds with $N=6$ and $F=44$, and hence, they are natural candidate models for ILSVRC mobile setting. Table~\ref{tab:cifar10_search_choose_i} reports the test error rates on CIFAR10, and we see that $I=3$ yields the best results. 

\subsection{L1 Regularization Constant $\lambda$}
\label{sec:l1_lambda_select}

\begin{table}[t]
    \centering
    \caption{Test error rates on CIFAR-10 by models found with different regularization constant $\lambda$.
    }
    \label{tab:cifar10_search_choose_lambda}
    \begin{tabular}{c|c}
    \hline
    Regularization Constant $\lambda$ & Average Lowest Error Rate \\
    \hline
    $0.1$ & 3.02  \\ 
    \textbf{0.001} & 2.88 \\  
    $0.00001$ & 3.13 \\ 
    \hline
    \end{tabular}
    
\end{table}

We choose the L1 regularization constant $\lambda$ of Eq.~\ref{eq:x_c_select_loss} to be 0.001 
from the range of $\{0.1, 0.001, 0.00001\}$, with the performances of the found models in Table~\ref{tab:cifar10_search_choose_lambda}.
High $\lambda$ means that the $L1$-regularization is highly valued, so that the shortcut selection is more sparse. However, strong regularization also prevents weak learners to fit their target loss gradient well. Since we mainly aim to select the most relevant shortcuts, and not to enforce the strict sparsity, we favor a small regularization constant.  

We also note that ~\citep{huang2017condensenet} has previously applied group Lasso to select filters in a DenseNet~\citep{densenet}. They apply a changing regularization constant $\lambda$ that gradually increases throughout the training. It will be interesting future improvement to select weak learners through dynamically changed regularization during weak learning.

\section{Search results on Penn Treebank (PTB)}
\label{sec:ptb_results}

\begin{table*}[t]
    \centering
    \caption{Comparison against state-of-the-art language modeling results on PTB. 
    We report \Petridish results in the format of ``best $|$ mean $\pm$ standard deviation'' from $10$ repetitions of the search with different random seeds.
    $^{*}$ From Table 2 in~\citep{randnas}.
    ${\dagger}$ \citep{randnas} report being unable to reproduce the DARTS results and this entry represents the results of DARTS (second order) as obtained via their deterministic implementation.
    $^{**}$ \citep{randnas} report being unable to reproduce ENAS results from original source code.
    $^{***}$ ENAS results as reproduced via DARTS source code.
    }
    \label{tab:ptb_search}
    \resizebox{\textwidth}{!}{
    \begin{tabular}{l|cccc}
    \hline
\multirow{ 2}{*}{\textbf{Method} }
        &  \textbf{\# params} 
        &  \textbf{Search } 
        &  \textbf{Test Error } \\
        &  (M)
        &  (GPU-Days)
        &  (perplexity)\\
\hline
Darts (first order)~\citep{Liu2018DARTSDA}$^{*}$
    & 23 &  1.5 &  57.6  \\
Darts (second order)~\citep{Liu2018DARTSDA}$^{*}$
    &  23 & 2 &  55.7  \\
Darts (second order)~\citep{Liu2018DARTSDA}$^{*}$ $^{\dagger}$
    &  23 & 2 & 55.9  \\
ENAS~\citep{Pham2018EfficientNA}$^{**}$
    & 24 & 0.5 & 56.3 \\
ENAS~\citep{Pham2018EfficientNA}$^{***}$
    & 24 & 0.5 & 58.6 \\
Random search baseline~\citep{randnas}$^{*}$
    &  23 & 2 & 59.4 \\ 
Random search WS~\citep{randnas}$^{*}$
    &  23 & 1.25 & 55.5 \\
\hline
\textbf{\Petridish} 
    & 23 & 1 & 55.85 $|$ 56.39$\pm$ 0.38 \\
\hline
    \end{tabular}
    }
\end{table*}

PTB \citep{ptb} has become a standard dataset in the NAS community for benchmarking NAS algorithms for RNNs. We apply \Petridish to search for the cell architectures of a recurrent neural network (RNN) \footnote{Note that for the case of architecture search of RNNs, cell-search and macro-search are equivalent.}. To keep the results as comparable as possible to most recent and well-performing work we keep the search space the same as used by DARTS~\citep{Liu2018DARTSDA} which in turn is also used byvery recent work~\citep{randnas}. There is a set of five primitives \texttt{\{sigmoid, relu, tanh, identity, none\}} that can be chosen amongst to decide connections between nodes in the cell. We modify the source code provided by~\cite{Liu2018DARTSDA} to implement \Petridish where we iteratively grow starting from a cell which contains only a single node \texttt{relu} connected to the incoming hidden activation and current input, until we have a total of $9$ nodes in the cell to match the size used in DARTS. At each stage of growth we train directly with an embedding size of $850$, $25$ epochs, $64$ batch size and a L1 weight of $10$ and select the candidate with the highest L1 weight value. We then add this candidate to the cell by removing the stop-gradient and stop-forward layers and replacing with regular connections. Table \ref{tab:ptb_search} shows a summary of the results. The rest of the parameters were kept the same as that used by~\cite{Liu2018DARTSDA}.

The final genotype obtained from the search procedure is then trained from scratch for $4500$ epochs, learning rate of $10$ and batch size $64$ to obtain final test perplexity reported below. We repeat the search procedure $8$ times with different random seeds and report the best and average test perplexity along with the standard deviation across search trials. Table \ref{tab:ptb_search} shows the results of running \Petridish on PTB. \Petridish obtains comparable results to DARTS, ENAS and Random Search WS. 

Note that since random search is essentially state-of-the-art search algorithm on PTB\footnote{As noted by \cite{randnas} current human-designed architecture by \cite{yang2018breaking} still beats the best NAS results albeit using a mixture-of-experts layer which is not in the search space used by DARTS, ENAS, and \Petridish to keep results comparable.}  we caution the community to not use PTB as a benchmark for comparing search algorithms for RNNs. The merits of any particular algorithm are difficult to compare at least on this particular dataset and task pairing. More research along the lines of \cite{nasbench} is needed on 1. whether the nature of the search space for RNNs specific to language modeling is particularly amenable to random search and or 2. whether it is the specific nature of RNNs by itself such that random search is competitive on any task which uses RNNs as the hypothesis space. We are presenting the results on PTB for the sake of completion since it has become one of the default benchmarks but ourselves don't derive any particular signal either way in spite of competitive performance.

\end{document}